\documentclass[sigconf]{acmart}
\renewcommand\footnotetextcopyrightpermission[1]{} 
\usepackage{epsfig}
\usepackage{graphicx}
\usepackage{amsmath}
\usepackage{amssymb}

\usepackage{booktabs}
\usepackage{color}
\usepackage{subfig} 
\usepackage{multirow} 

\usepackage{caption}
\usepackage{float}
\usepackage{array}

\def\BibTeX{{\rm B\kern-.05em{\sc i\kern-.025em b}\kern-.08emT\kern-.1667em\lower.7ex\hbox{E}\kern-.125emX}}

\copyrightyear{2019}
\acmYear{2019}
\setcopyright{acmlicensed}
\acmConference[MM '19]{MM '19: In 2019 ACM Multimedia Conference}{October 21--25, 2019}{Nice, France}
\acmBooktitle{2019 ACM Multimedia Conference (MM '19), October 21--25, 2019, Nice, France}
\acmPrice{15.00}
\acmDOI{10.1145/1122445.1122456}
\acmISBN{978-1-4503-9999-9/18/06}

\acmSubmissionID{678}

\begin{document}

%
\title{A Novel Teacher-Student Learning Framework\\
 For Occluded Person Re-Identification  }

%
%
%

\author{Jiaxuan Zhuo}
\affiliation{%
  \institution{School of Electronics and Information Technology}
  \city{Sun Yat-sen University}
  \country{China}
}
\email{zhuojx5@mail2.sysu.edu.cn}

\author{Jianhuang Lai}
\authornotemark[1]
\affiliation{%
  \institution{School of Data and Computer Science}
  \city{Sun Yat-sen University}
  \country{China}
}
\email{stsljh@mail.sysu.edu.cn}

\author{Peijia Chen}
\affiliation{%
  \institution{School of Data and Computer Science}
  \city{Sun Yat-sen University}
  \country{China}
}
\email{chenpj8@mail2.sysu.edu.cn}

%
%
%
%
%

%

%
\begin{abstract}
   Person re-identification (re-id) has made great progress in recent years, but occlusion is still a challenging problem which significantly degenerates the identification performance. In this paper, we design a teacher-student learning framework to learn an occlusion-robust model from the full-body person domain to the occluded person domain. Notably, the teacher network only uses large-scale full-body person data to simulate the learning process of occluded person re-id. Based on the teacher network, the student network then trains a better model by using inadequate real-world occluded person data. In order to transfer more knowledge from the teacher network to the student network, we equip the proposed framework with a co-saliency network and a cross-domain simulator. The co-saliency network extracts the backbone features, and two separated collaborative branches are followed by the backbone. One branch is a classification branch for identity recognition and the other is a co-saliency branch for guiding the network to highlight meaningful parts without any manual annotation. The cross-domain simulator generates artificial occlusions on full-body person data under a growing probability so that the teacher network could train a cross-domain model by observing more and more occluded cases. Experiments on four occluded person re-id benchmarks show that our method outperforms other state-of-the-art methods.
\end{abstract}


%

%

%
\maketitle

\section{Introduction}

With the explosive growth of video surveillance systems, person re-identification (re-id) is an increasingly significant task for searching specific persons in surveillance, e.g., criminals, children and other missing persons, and has been widely researched and developed. However, in real-world applications of person re-id, it would easily occur that the target pedestrian is partially occluded by some dynamic or static obstacles, such as other persons, cars, signposts, pillars, etc, especially in crowded places. In the case of occlusions, existing person re-id algorithms no longer perform well because they assume that all detected targets are of full-body persons.

\begin{figure}[!tp]
	\begin{center}		
		\includegraphics[width=0.44\textwidth,height=0.22\textheight]{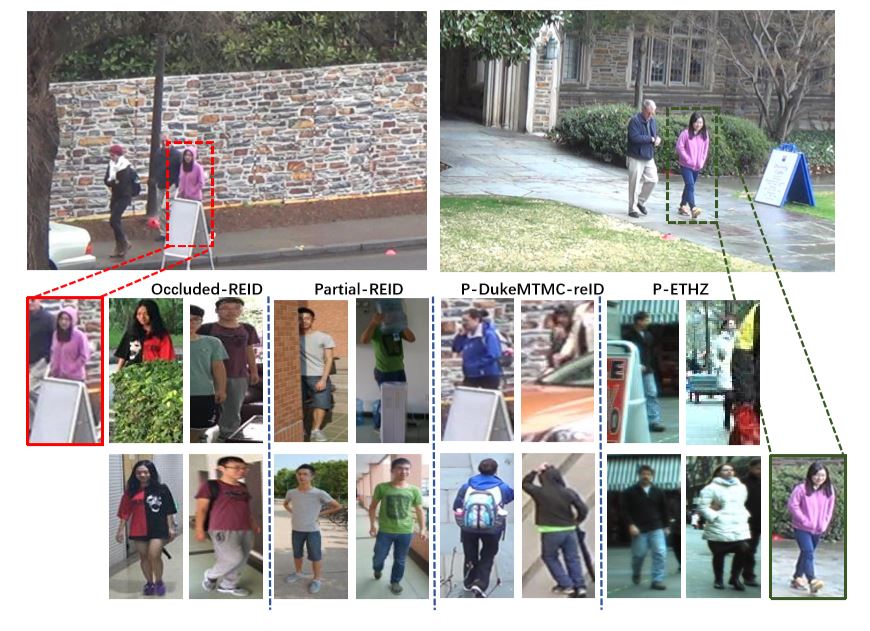}
		\caption{Scenario of occluded person re-id and samples from occluded person re-id datasets. For a person with occlusions detected from one camera, occluded person re-id task aims to search for the same identity from the pedestrian database or other non-overlapping cameras.}
        \label{Fig:pic1}
	\end{center}
\end{figure}

\begin{figure}[t]
	\begin{center}		
		\includegraphics[width=0.44\textwidth]{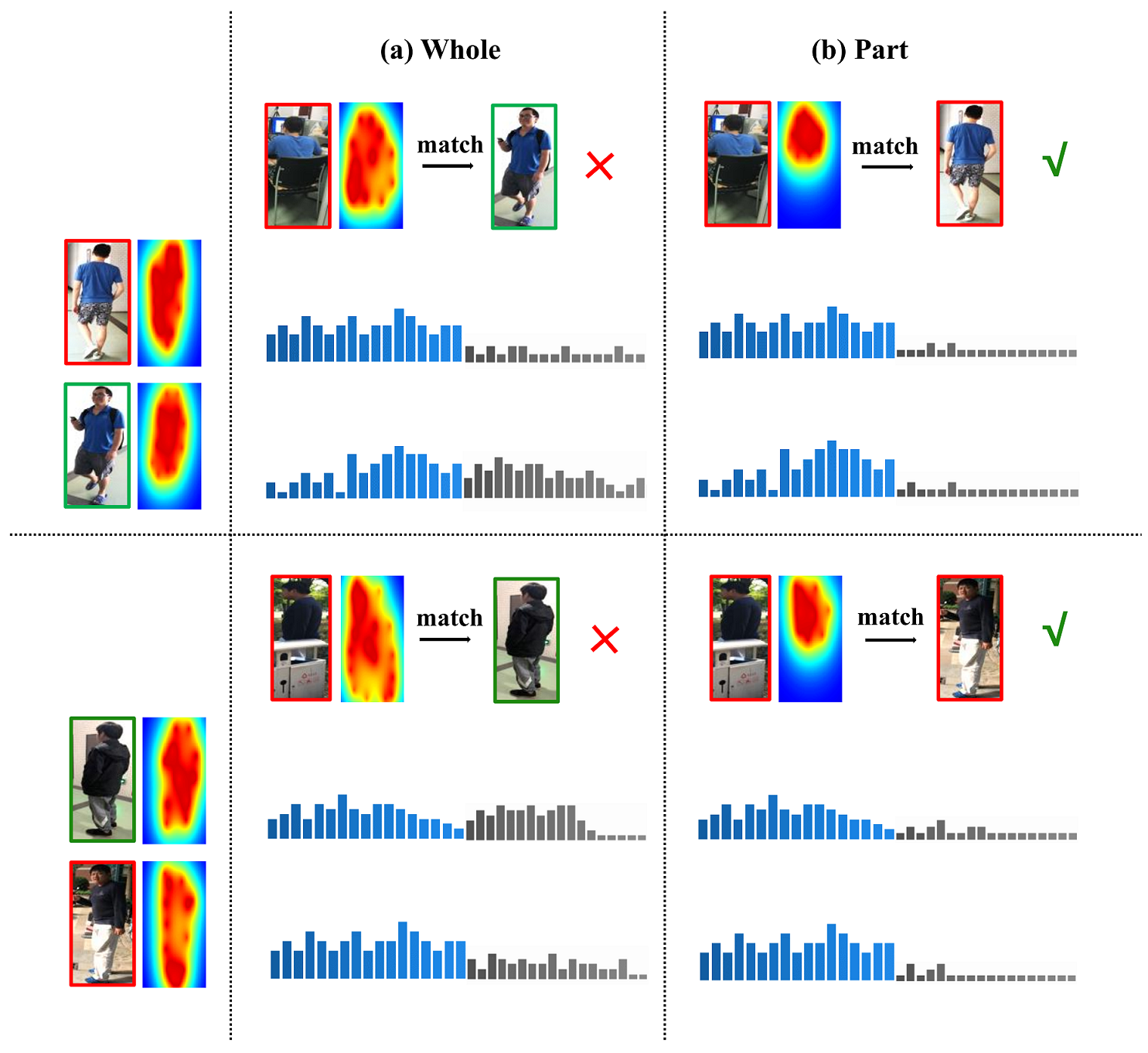}
		\caption{Motivation of our proposal. The histogram represents the similarity between queries and galleries. Red boxes mean the same one, while green boxes mean different ones. Saliency maps beside images show the regions where the network focus. (a) When extracting features over the whole image, it would easily cause mismatching. (b) When extracting features focusing on the essential parts, it is more likely to match correctly.}
		\label{Fig:pic2}
	\end{center}
\end{figure}

Occluded person re-id task \cite{Zheng2016Partial,Zhuo2018Occluded,DBLP:conf/cvpr/HeLLS18,DBLP:journals/corr/abs-1810-06996} is to re-identify the same person given a detected target person with occlusions (Figure \ref{Fig:pic1}). Different from the full-body person domain, the occluded person domain contains several detected target persons covered with various partial occlusions, making it hard to get the correct matching in occluded person re-id task. The challenges of the occluded person re-id are listed as follows. First, following the ordinary practice in person re-id, features extracted by traditional filters or convolutional neural networks (CNNs) over the whole images would be easily corrupted by the occluded regions. As shown in Figure \ref{Fig:pic2}(a), it is more likely to cause misleading matching due to the interference from occlusions. Second, some works \cite{Zhou_2018_ECCV,Zhang_2018_ECCV,DBLP:conf/eccv/ZhuWLWYH18} suggested to make use of the pedestrian detection to obtain tight bounding boxes of person body parts so as to extract effective features. However, occlusions have unknown spatial positions, sizes, shapes and colors in different cases, and are even incomplete objects. Hence it is most unlikely to train a toolbox to accurately separate pedestrians and occlusions. Furthermore, even if the bounding box of only person parts can be detected, as \cite{Zheng2016Partial}, there are also some parts of the person body that cannot be included or covered by residuary occlusions because the boundary between the target person and occlusions is always not a flat edge. In short, it is difficult to delicately eliminate occlusions in this task. Last but not least, one of the most serious drawbacks is that, there are insufficient occluded person data to learn a robust model for occluded person re-id while most of existing public large-scale datasets belong to the full-body person domain, which is also a problem need to be solved.

In order to address occluded person re-id, we consider training a model which pays more attention to person body parts rather than the whole image. Based on the above consideration, we require the model to capture the essential clues mainly from  target persons and ignore other useless information from occlusions or backgrounds. If the essential parts play a major role in features, incorrect matching could be reduced, as shown in Figure \ref{Fig:pic2}(b). But it is difficult to train such a robust model without adequate occluded person data. Therefore, we consider taking advantage of large-scale full-body person data to improve the learning of occluded person data.

Considering the above issues, we propose a teacher-student learning framework consisting of the "teacher" stage and the "student" stage. In the "teacher" stage, we only make use of large-scale full-body person data to simulate the occluded person re-id task. The "teacher" provides the "student" with a basic scheme to address the real-world occluded person re-id problem. We design a co-saliency network as the teacher network together with a cross-domain simulator. The co-saliency network extracts the backbone features following by two branches, a classification branch and a co-saliency branch. Two branches both feedback to the shared backbone, though they finish different tasks: 1) the classification branch acts as an identity classifier 2) while the co-saliency branch aims to separate pixel-wise predominant regions that is person body parts from the rest of the image. The former enables the network to identify different persons and the latter helps the network capture essential clues by highlighting salient person body parts. Particularly, when training the co-saliency branch, the ground truth comes from the masks predicted by an existing salient object detector rather than by manual annotation.
To offer the teacher network simulated occluded material, the cross-domain simulator is designed to transform full-body person data to simulated occluded person data by generating various artificial occlusions over full-body person images with a growing probability during training. As the iteration increasing, the "teacher" learns a more occlusion-robust model by observing more and more simulated occluded person images. The "student" then inherits a basic model from the "teacher" and trains it on real-world occluded person data. Besides, the co-saliency branch of the "teacher" is used to predict the masks of occluded person images, which offer the better ground truth to the "student".

Consequently, our framework combines simulation teaching of the "teacher" and practical learning of the "student" to reach better performance in occluded person re-id, advantages of which are as follows: 1) with the help of the teacher-student learning, we break through the restriction of inadequate occluded person data. 2) The "teacher" provides an effective network paying more attention to person body parts, from which the "student" gets a better model to deal with the harmful interference of occlusions.

In summary, this paper makes three main contributions.
\begin{itemize}
  \item We develop a novel two-stage teacher-student learning framework to solve the challenges of occluded person re-id by building a bridge across the full-body person domain to the occluded person domain.
  \item We propose a co-saliency network together with a cross-domain simulator, which trains an occlusion-robust model paying attention to person body parts.
  \item Furthermore, the co-saliency branch achieves better performance for occluded person detection than other salient object detectors, and our proposal shows superiority against of the state-of-the-art occluded person re-id methods.
\end{itemize}

\begin{figure*}
	\begin{center}		
		\includegraphics[width=0.9\textwidth]{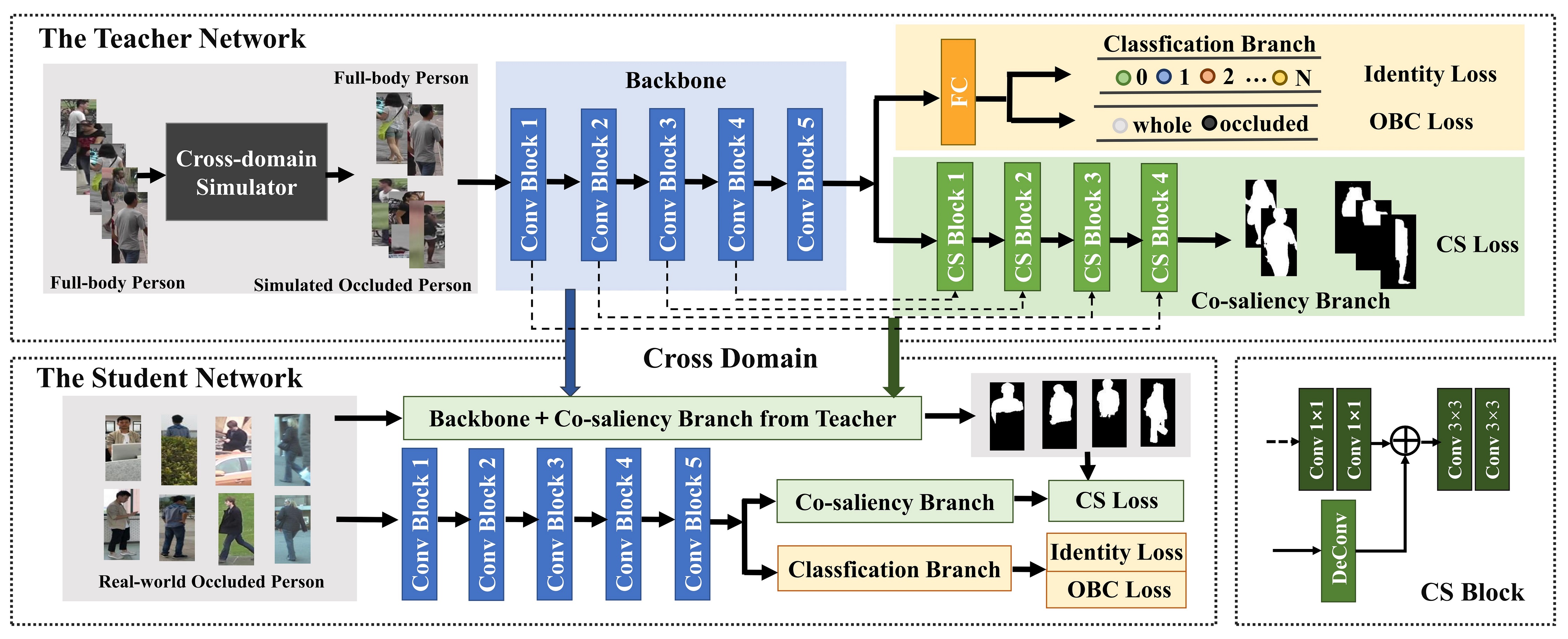}
		\caption{Overview of our proposed framework. The teacher network learns a basic occlusion-robust model only using large-scale full-body person data, which adopts a co-saliency network with a cross-domain simulator to simulate the occluded person re-id task. The student network then practices on real-world occluded person data based on the teacher network.}
		\label{Fig:pic4_1}
	\end{center}
\end{figure*}

\section{Related Work}
\noindent{\bf Person re-identification.}\ Recently, person re-id has made great progress, which mainly consists of two components, a feature extractor for describing images \cite{Li2017Learning,xiao2016learning,Cheng2016Person,Sun2018Beyond,Qian2017The,DBLP:conf/iccv/LiuZTSSYYW17,chang2018multi,DBLP:conf/cvpr/KalayehBGKS18} and a similarity metric for distance learning \cite{Cheng2016Person,Hermans2017In,Chen2017Beyond,Varior2016Gated,DBLP:conf/cvpr/HeLLS18}. Previous works that focus on feature extractors are divided into two major groups, the local ones \cite{liao2015person,Li2017Learning} and the global ones \cite{xiao2016learning}. For example, Sun et al. \cite{Sun2018Beyond} proposed a network named Part-based Convolutional Baseline (PCB), which could output a convolutional descriptor consisting of several part-level features. Yu et al. \cite{Qian2017The} argued that both high-level and mid-level features were relevant for Cross-Domain Instance Matching (CDIM). To extract more discriminative features, Chang et al. \cite{chang2018multi} proposed Multi-Level Factorisation Net (MLFN), a novel network architecture that factorised the visual appearance of a person into latent discriminative factors at multiple semantic levels without manual annotation. Meanwhile, some other works that focused on similarity metric also have grown rapidly in recent years. Varior et al. \cite{Varior2016Gated} proposed a gated Siamese Convolutional Neural Network (gated S-CNN) to distinguish positive pairs from hard negative pairs by finer local patterns. Cheng et al. \cite{Cheng2016Person} presented a novel Multi-Channel Parts-based Convolutional Neural Network model with improved triplet loss function, which could pull the instances of the same person closer and push the instances of different persons farther from each other in the learned feature space. Different from most person re-id methods training on full-body person datasets, approaches to solve the occluded person re-id problem tend to move along at a slow.

 \noindent{\bf Occluded person re-id.}\ As person re-id are applied in real-world applications more frequently, some practical challenges in person re-id have gained widely concern, one of which is occlusion. However, there are a few methods \cite{Zheng2016Partial,DBLP:conf/cvpr/HeLLS18,DBLP:journals/corr/abs-1810-06996} considering how to cope with the occluded person re-id problem. Zheng et al. \cite{Zheng2016Partial} proposed a local patch-level matching model namely Ambiguity-sensitive Matching Classifier (AMC) as well as a global-to-local matching model namely Sliding Window Matching (SWM), which was the earliest work to overcome the occluded person re-id problem. However, the computation cost of AMC-SWM was expensive because it extracted multi-patch features from a large number of grid patches respectively. In recent works, He et al. \cite{DBLP:conf/cvpr/HeLLS18} introduced a method called Deep Spatial feature Reconstruction (DSR), which was flexible to calculate the similarity of feature blocks of different sizes. Fan et al. \cite{DBLP:journals/corr/abs-1810-06996} then proposed an unsupervised method called Spatial-Channel Parallelism Network (SCPNet), which effectively used the local features to leverage the global features.


As mentioned above, most of previous works for occluded person re-id focus on extracting both local-patch features and global-image features so as to acquire more information. However, the computation cost and the misalignment problem of local patches would depress the efficiency and effect of the task. Different from utilizing local patches or features, our method can capture the essential clues from the whole image by paying more attention to person body parts. And a teacher-student learning framework is designed to achieve better robustness against occlusions in practice.

\section{Proposed Method}

In this work, we present a teacher-student learning framework for occluded person re-id, which is a two-stage training across from the full-body person domain to the occluded person domain. In this section we detail the implementation of the proposed framework.

\subsection{Overview}
We focus on the occluded person re-id problem suffered by limited occluded training data and unreliable features from occlusions, and address this key problem through a teacher-student learning framework (Figure \ref{Fig:pic4_1}). This framework contains two stages, the "teacher" stage for simulation teaching and the "student" stage for practical learning, as introduced in the following.

In the "teacher" stage, we make use of large-scale full-body person data to simulate the learning process of occluded person re-id, which we call simulation teaching. To reach it, we design a co-saliency network together with a cross-domain simulator. On the one hand, the co-saliency network extracts the backbone features and develops two separated branches, one of which is the classification branch for identity recognition and the other is the co-saliency branch for salient person detection. The ground truth of the co-saliency branch comes from a existing salient object detector. Two branches constrain the backbone feature extractor to focus on person body parts in order to extract occlusion-robust features. On the other hand, the cross-domain simulator is used to transform full-body person data to simulated occluded person data with a growing probability during training. Benefiting from the cross-domain simulator, the teacher network could extract discriminative and occlusion-robust features by training on full-body person data and simulated occluded person data jointly. After a period of learning, the teacher network would learn a basic occlusion-robust model by observing various cases of simulated occlusions.

Relying on the "teacher" stage, the "student" carries forward the basic model and trains it on real-world occluded person data, which is called as practical learning. It is worth mentioning that the co-saliency branch could improve the performance of salient occluded person detection. We thus use the co-saliency branch from the "teacher" stage to generate the ground truth for the "student" stage instead of the initial salient object detector.

In general, our framework implements the occlusion-robust learning across from the full-body person domain to the occluded person domain. Experiments in Section 4 would show the effectiveness of our framework in the supervised and unsupervised occluded person re-id tasks.

\subsection{Co-saliency Network }
As pointed out in Section 1, the main drawback of occluded person re-id arises from unreliable features of occlusions, which easily causes misleading matching (Figure \ref{Fig:pic2}(a)). To avoid undesirable effects of occlusions ,we consider to extract occlusion-robust features by paying attention to person body parts, so we propose a co-saliency network.

The co-saliency network consists of one backbone and two separated collaborative branches. The backbone firstly conducts as a feature extractor to exploit rich features by several convolution layers. All convolution layers in the backbone are divided into five convolution blocks, in which the resolution of feature maps is the half of the preceding one. Two branches, namely the classification branch and the co-saliency branch, are then constructed on the shared backbone for guiding the feature extraction. The classification branch is formed by a fully-connected layer and an identity classification loss function, which enforces the backbone to learn discriminative features to minimize the classification error. Another branch, the co-saliency one composed of four co-saliency (CS) blocks and a co-saliency loss (CS loss) function is serving for image-to-image translation that is salient person detection. Each CS block utilizes a deconvolution layer with bilinear interpolation to upsample the preceding feature maps by two times and parallel stacks two $1\times1$ convolution layers to transform the intermediate feature maps from the backbone blocks into ones with the uniform channels. After an pixel-wise summation operation for fusing these two output feature maps,  two $3\times3$ convolution layers are used to convert the fused feature maps into compatible contextual information. Lastly, the CS loss function calculates the similarity error between the predicted saliency masks and the ground truth by the pixel-level classification. The co-saliency branch achieves finer saliency detection by combining multi-level features from deep and shallow side-output layers, which optimises the backbone features more concerned with salient person body regions.

In the co-saliency network, the backbone and the classification branch act as the feature extractor $f(\bullet)$ and the identity classifier $g(\bullet)$, respectively. Suppose that we have $D$ domains, each of which has $N_i$ images of $C_i$ identities and $i=\{1,2,\ldots,D\}$. Let $I_i^{(j)}$  denote the $j_{th}$ image of the $i_{th}$ domain while $C_i=\{c_i^{(1)}, c_i^{(2)}, \ldots, c_i^{(N_i)}\}$ and $S_i=\{s_i^{(1)}, s_i^{(2)}, \ldots, s_i^{(N_i)}\}$ are respectively the identity labels and the saliency ground truth. All samples in domain $D_i$ are expressed as $(I_i^{(j)}, c_i^{(j)}, s_i^{(j)})_{j=1}^{N_i}$. Thus, $f(\bullet)$ and $g(\bullet)$ can be optimized by the formula
\begin{equation}
\underset{f,g}{\arg\min}\sum_{j=1}^{N_i}\mathcal{L}(g(f(I_{i}^{(j)})), c_i^{(j)}) , \\
\end{equation}
where $\mathcal{L}$ is the softmax loss function that is equal to the cross-entropy between the predicted result and the ground truth. The co-saliency branch is set as a salient person detector $h(\bullet)$ to localize the most conspicuous regions, which can be optimized by the formula
\begin{equation}
\underset{f,h}{\arg\min}\sum_{j=1}^{N_i}\sum_{q=0}^{W_i}\sum_{p=0}^{H_i}\mathcal{L}(h(f(I_{i,(p,q)}^{(j)})), s_{i,(p,q)}^{(j)}) , \\
\end{equation}
where $( p, q )$ is the position of a pixel in a ${H_i}\times{W_i}$ image. Combined the backbone with two branches, the loss function of the co-saliency network can be represented as
\begin{equation}
\begin{split}
\mathcal{L} = \alpha\mathcal{L}^C(I_i, C_i)+(1-\alpha)\mathcal{L}^S(I_i, S_i) , \\
\end{split}
\end{equation}
where $\mathcal{L}^C(I_i, C_i)$ and $\mathcal{L}^S(I_i, S_i)$ represent the loss functions of the classification branch and co-saliency branch respectively and $\alpha\in(0,1)$ is a hyper-parameter to balance two separated branches. Since the classification branch play a major role in occluded person re-id, it is reasonable to set $\alpha\geq0.5$ in the loss function.

It can be seen that the classification branch and the co-saliency branch establish an interacting relationship by sharing the backbone features from $f(\bullet)$. Furthermore, the two branches also promote with each other because of the consistent goal aiming to obtain reliable features through focusing on person body parts. They help each other forward, just like two collaborators in a work. The co-saliency branch brings an impressive improvement for the classification branch by encoding the location information of person body parts into the shared features. Meanwhile, The classification branch transmits the meaningful semantic information of the pedestrians to the co-saliency branch through the shared features. Fact proved that the co-saliency network enhances features occlusion-robust capability owing to the beneficial interaction between the co-saliency branch and the classification branch.

\subsection{Cross-domain Simulator}

To mitigate the limitation of inadequate occluded person data, we develop a teacher-student learning framework, which employs large-scale full-body person data to simulate the training of occluded person data. Nevertheless, the performance would be significantly degraded when directly conducting the training across different domains, from the full-body person domain to the occluded person domain. The key to this problem is to narrow the difference between these two domains. Therefore, we design a cross-domain simulator, which constructs a gradual domain migration from full-body person data to simulated occluded person data. The cross-domain simulator works on the data loading stage of each iteration in the co-saliency network. It firstly remains the identity labels unchanged and then selects some of full-body person images and saliency ground truth at a setting probability to cover with various artificial occlusions. Specifically, the probability is growing with the increase of iterations so that simulated occluded person data can get more involved in training. Moreover, the selected samples would be marked with a new label called occluded/non-occluded binary classification (OBC) label to present the occluded one while the rest would be marked with the non-occluded one. The OBC labels are used for the occluded/non-occluded binary classification loss function so that we could integrate the identity classification loss and the OBC loss into the classification branch. The procedure of the training with the cross-domain simulator is shown in Algorithm 1.

The OBC classifier $b(\bullet)$ aims to determine whether a sample is from the full-body person domain or the occluded person domain, which can be optimized by the formula
\begin{equation}
\underset{f,b}{\arg\min}\sum_{j=1}^{N_i}\mathcal{L}(b(f(I_{i}^{(j)})), o_i^{(j)}) , \\
\end{equation}
where $O_i=\{o_i^1,o_i^2,\ldots,o_i^{N_i}\}, O_i\in\{0, 1\}$ is the OBC labels and $0, 1$ denote the non-occluded one and the occluded one, respectively.

Inspired by the cross-domain simulator, we combine the identity classification loss and the OBC loss as the multi-task loss in the classification branch as \cite{Zhuo2018Occluded}. The loss function of the classification branch is given by
\begin{equation}
\begin{split}
\mathcal{L}^{M}(I_i, C_i, O_i) = \beta\mathcal{L}^C(I_i, C_i)+(1-\beta)\mathcal{L}^O(I_i, O_i) ,
\end{split}
\end{equation}
where $\mathcal{L}^O(I_i, O_i)$ represents the OBC loss function. $\beta\in(0,1)$ is a hyper-parameter which balances the proportion of two classifiers in the classification branch, which is always set to more than 0.5. Thereby, we get the final loss function as
\begin{equation}
\begin{split}
\mathcal{L}=\sum_{i=1}^{D}&(\alpha\mathcal{L}^M(I_i, C_i, O_i)+(1-\alpha)\mathcal{L}^S(I_i, S_i)) , \\
\end{split}
\end{equation}
where $D$ is 2 because there are the full-body person domain and the occluded person domain using in our framework.

In general, there are three advantages using the cross-domain simulator: 1) The network can simultaneously optimise features discriminative capability by training on full-body person data and occlusion-robust capability by training on simulated occluded person data. Since full-body person data are more likely to be transformed to simulated occluded person data with the increase of iteration, the network gradually enhances the robustness against occlusions by observing more simulated occluded cases. 2) The OBC loss could be jointed into the network, which encodes the prior information whether the person is occluded or not into the framework. 3) The treatment of saliency ground truth could produce a variety of training pairs to achieve data augmentation so that the co-saliency branch improves the capacity for detecting salient occluded persons.

\begin{table}[t]
\footnotesize
	\begin{tabular}{l}
		\hline\noalign{\smallskip}
		\textbf{Algorithm 1:} Training with the cross-domain simulator \\
		\hline\noalign{\smallskip} 	
		\textbf{Input:} Full-body person images $I$ ($N$ images), \\
            \quad\quad\quad identity labels $C$ and saliency masks $S$ \\
            \quad\quad\quad Max epoch of training $epoch_{max}$ \\
            \quad\quad\quad The growing probability of occlusions $p$ \\
		\textbf{Output:} The occlusion-robust basic model \\
        \textbf{Initial:} $p \leftarrow 0$, $epoch \leftarrow 0$ \\
        1:\,\,\textbf{while} $epoch < epoch_{max}$ \textbf{do:} \\
		2:\,\,\quad if $I \in$ select $p \times N$ images randomly: \\
        3:\,\,\quad\quad put $(I,C,S)$ into the cross-domain simulator \\
		4:\,\,\quad\quad $I_s \leftarrow$ paste a background patch to $I$ randomly \\
        5:\,\,\quad\quad $S_s \leftarrow$ paste a black patch to $S$ in the same position \\
        6:\,\,\quad\quad $O_s \leftarrow$ label 1 (occluded person) \\
        7:\,\,\quad\quad $C_s \leftarrow C$ remain unchanged \\
        8:\,\,\quad else: \\
        9:\,\,\quad\quad ${O_r} \leftarrow$ label 0 (non-occluded person) \\
        10:\quad\quad $\{I_r,C_r,S_r\}$ $\leftarrow$ $\{I,C,S\}$ remain unchanged \\
        11:\quad train the model $\leftarrow$ combine $\{I_r,C_r,S_r,O_r\}$ and $\{I_s,C_s,S_s,O_s\}$ \\
        12:\quad $epoch \leftarrow epoch + 1$ \\
        13:\quad $p \leftarrow epoch/epoch_{max}$ \\
        14:\,\textbf{end while}\\
		\hline
	\end{tabular}
\end{table}

\section{Experiments}

\subsection{Datasets}
We conduct our experiments on four occluded person re-id datasets, Occluded-REID, Partial-REID, P-DukeMTMC-reID, P-ETHZ and a large-scale full-body person dataset, MARS, which are introduced as follows.

\textbf{Occluded-REID} \cite{Zhuo2018Occluded} is an occluded person dataset captured by mobile cameras with different viewpoints and backgrounds. There are two folders, the occluded one and the whole one, including 2,000 images of 200 identities. Each identity has 5 full-body person images and 5 occluded person images with different types of occlusions.

\textbf{Partial-REID} \cite{Zheng2016Partial} is the first partial person re-id dataset including 900 images of 60 persons. Each person has 5 full-body person images, 5 partial person images and 5 occluded person images with various occlusions. All the images were collected at a campus.

\textbf{P-DukeMTMC-reID} is a subset of DukeMTMC \cite{ristani2016performance} captured by multiple cameras, which recorded outdoors on the campus. There are 24,143 images of 1,299 identities and each identity has both full-body person images and occluded person images .

\textbf{P-ETHZ} is a subset of pedestrian dataset ETHZ \cite{ess2008mobile}. This dataset does have considerable illumination variance, scale variance and occlusion. Following \cite{Zhuo2018Occluded}, we use 3,897 images of 85 identities with both full-body person images and occluded person images.

\textbf{MARS} \cite{zheng2016mars}, an extension of Market-1501 \cite{DBLP:conf/iccv/ZhengSTWWT15}, is the first large-scale video based person re-id dataset. It consists of 1,191,003 full-body person images of 1,261 different pedestrians from 20,478 video sequences and 1,191,003 bounding boxes captured by 6 cameras.

\begin{table*}[t]
\caption{Evaluation of key components in the proposed framework. "T$^\dagger$" and "S$^\dagger$" mean the "teacher" stage and the "student" stage. The representations of key components are as follows: "S": the co-saliency branch, "C": the classification branch, "D": the cross-domain simulator, "O": the OBC loss. The top three results are highlighted in \textcolor{red}{red}, \textcolor{blue}{blue} and \textcolor{green}{green}, respectively.}
\label{Tab:table1}
\footnotesize
\tabcolsep4pt 
\renewcommand\arraystretch{1.25} 
\begin{center}
\begin{tabular}{c|l|cccc|cccc|cccc|cccc}
\hline
 & Dataset & \multicolumn{4}{c|}{Occluded-REID} & \multicolumn{4}{c|}{Partial-REID} & \multicolumn{4}{c|}{P-DukeMTMC} & \multicolumn{4}{c}{P-ETHZ} \\ \hline
 & Method & r=1 & r=2 & r=5 & mAP & r=1 & r=2 & r=5 & mAP & r=1 & r=2 & r=5 & mAP & r=1 & r=2 & r=5 & mAP \\ \hline
\multirow{6}{*}{w/o-S$^\dagger$} & w/o-T$^\dagger$ & 3.60 &6.60  &14.60  &7.15  &10.67  &16.67  &30.67  &16.48  &1.10  &1.72  &3.15  &2.02  &25.48  &37.86  &52.86  &32.49  \\
 & T$^\dagger$:S & 11.30 & 16.40 & 26.10 & 15.73 &16.00  &27.67  &43.33  &23.35  & 2.15 &3.41  &5.62  &3.41  & 26.19 & 33.80 &44.29  &31.17  \\
 & T$^\dagger$:C & 46.75 & 57.00 & 66.75 & 51.57 & 58.33 &67.47  &76.67  &63.05  &15.47  &20.69  &28.19  &18.90  & 27.38 &\textcolor{green}{40.48}  & \textcolor{green}{48.81} & 33.14 \\
 & T$^\dagger$:C+S &  \textcolor{green}{49.75}& \textcolor{green}{60.25} &\textcolor{blue}{70.50}  &\textcolor{green}{54.66}  &\textcolor{green}{62.50}  &\textcolor{green}{69.17}  &\textcolor{green}{85.00}  &\textcolor{green}{66.98}  &\textcolor{green}{17.76}  &\textcolor{green}{23.09}  &\textcolor{green}{31.36}  &\textcolor{green}{21.28}  & \textcolor{green}{28.10} & 37.62 & \textcolor{red}{55.48} & \textcolor{green}{34.59} \\
 & T$^\dagger$:C+S+D &  \textcolor{blue}{53.00}& \textcolor{blue}{61.50} &\textcolor{green}{70.25}  &\textcolor{blue}{57.24}  &\textcolor{blue}{67.50} & \textcolor{red}{80.00} & \textcolor{red}{86.67} & \textcolor{blue}{72.05} & \textcolor{blue}{18.35} &\textcolor{blue}{24.13}  &\textcolor{blue}{31.99}  &\textcolor{blue}{22.00}  &\textcolor{blue}{31.90}  & \textcolor{red}{41.43}  &\textcolor{blue}{53.81}  & \textcolor{blue}{37.41}  \\
 & T$^\dagger$:C+S+D+O(Ours) & \textcolor{red}{55.00} & \textcolor{red}{64.50} & \textcolor{red}{77.25}  & \textcolor{red}{59.84} & \textcolor{red}{69.17}  & \textcolor{blue}{76.67}  & \textcolor{blue}{85.83} & \textcolor{red}{73.11} & \textcolor{red}{18.80} & \textcolor{red}{24.23} & \textcolor{red}{32.21} & \textcolor{red}{22.37} & \textcolor{red}{33.33} & \textcolor{blue}{40.48} & 45.23 & \textcolor{red}{37.44}  \\ \hline
\multirow{6}{*}{w/-S$^\dagger$} & w/o-T$^\dagger$ & 24.20 & 32.80 & 48.40 & 30.22 & 25.67 & 36.67 & 56.67 & 32.98 &  36.15& 43.79 &54.70  &40.67  &30.95  &41.43  &53.33  &36.83  \\
 & T$^\dagger$:S & 63.79 & 75.60 & 86.00 & 68.85 & 60.00 & 72.67 & 88.33 & 65.91 & 40.44 & 49.54 & 59.93 & 45.14 & 45.39 & 58.10 & 71.59 & 51.74 \\
 & T$^\dagger$:C & 62.70 & 72.30 & 83.30 & 67.34 & 66.00 & 78.00 & 87.99 & 71.16 & 42.58 & 51.72 & 63.75 & 42.94 & 48.33 & 61.67 & \textcolor{green}{79.05} & 55.16 \\
 & T$^\dagger$:C+S & \textcolor{green}{68.50} & \textcolor{green}{78.79} & \textcolor{green}{86.90} & \textcolor{green}{72.77} & \textcolor{green}{72.67} & \textcolor{green}{81.99} & \textcolor{green}{90.67} & \textcolor{green}{76.70} & \textcolor{green}{47.27} & \textcolor{green}{55.36} & \textcolor{green}{65.20} & \textcolor{green}{51.49} & \textcolor{green}{54.28} & \textcolor{green}{67.62} & 75.24 & \textcolor{green}{59.91} \\
 & T$^\dagger$:C+S+D & \textcolor{blue}{72.40} & \textcolor{blue}{83.00} & \textcolor{blue}{90.80} & \textcolor{blue}{76.59} &  \textcolor{blue}{80.67} & \textcolor{blue}{90.67} & \textcolor{blue}{94.99} & \textcolor{blue}{84.06}  &  \textcolor{blue}{50.11} & \textcolor{red}{58.83} & \textcolor{blue}{68.40} & \textcolor{blue}{54.36} & \textcolor{blue}{60.48} & \textcolor{blue}{74.28} & \textcolor{blue}{83.81} & \textcolor{blue}{65.90} \\
 & T$^\dagger$:C+S+D+O(Ours) & \textcolor{red}{73.67} & \textcolor{red}{84.40} & \textcolor{red}{92.87}  & \textcolor{red}{77.89} & \textcolor{red}{82.67} & \textcolor{red}{91.33} & \textcolor{red}{97.00} & \textcolor{red}{85.87} & \textcolor{red}{51.42} & \textcolor{blue}{58.52} & \textcolor{red}{69.72} & \textcolor{red}{55.60} & \textcolor{red}{62.86} & \textcolor{red}{75.71} & \textcolor{red}{85.24} & \textcolor{red}{68.05} \\
\hline
\end{tabular}
\end{center}
\end{table*}

\subsection{Implementation Details}

 \noindent\textbf{Model.} We choose ResNet-50 \cite{he2016deep} as our feature backbone and our baseline is the backbone with a classification branch. Euclidean distance is used for similarity metric.

 \noindent\textbf{Optimization.} We implement our proposed method based on Pytorch \cite{paszke2017automatic}. The network is trained using the adaptive moment estimation optimizer (Adam \cite{DBLP:journals/corr/KingmaB14}). The learning rate of the backbone is set to 1$\times$10$^{-5}$ and two branches to 2$\times$10$^{-4}$. The hyper-parameter $\alpha$ and $\beta$ are both set to 0.8. We train the model on a single GPU with the batch size 8 for 50K iterations.

 \noindent\textbf{Data arrangement.} We take occluded person images as the probes and full-body person images as the galleries. The teacher network is only trained on the full-body person dataset, MARS. The student network is trained on half of an occluded person dataset and we test the rest when conducting the supervised experiments. While in unsupervised experiments, the whole occluded person dataset is used for testing without training in the "student" stage. The input images are resized to 240 $\times$ 240 and randomly cropped to 224 $\times$ 224 for training. For a fair comparison, the other contrast experiments use the same configuration as the above.

 \noindent\textbf{Evaluation.} The evaluations of the methods are mainly Cumulative Match Characteristic (CMC) \cite{gray2007evaluating} and mean Average Precision (mAP) \cite{zheng2015scalable} in person re-id. Besides, we use Precision, Recall and F-measure score ($\beta^2$=0.3) \cite{DBLP:conf/cvpr/2009} to evaluate the effectiveness of the co-saliency branch.

\subsection{Ablation Analysis}

\begin{figure*}[t]
	\begin{center}		
		\subfloat[Occluded-REID]{\includegraphics[width=0.24\textwidth]{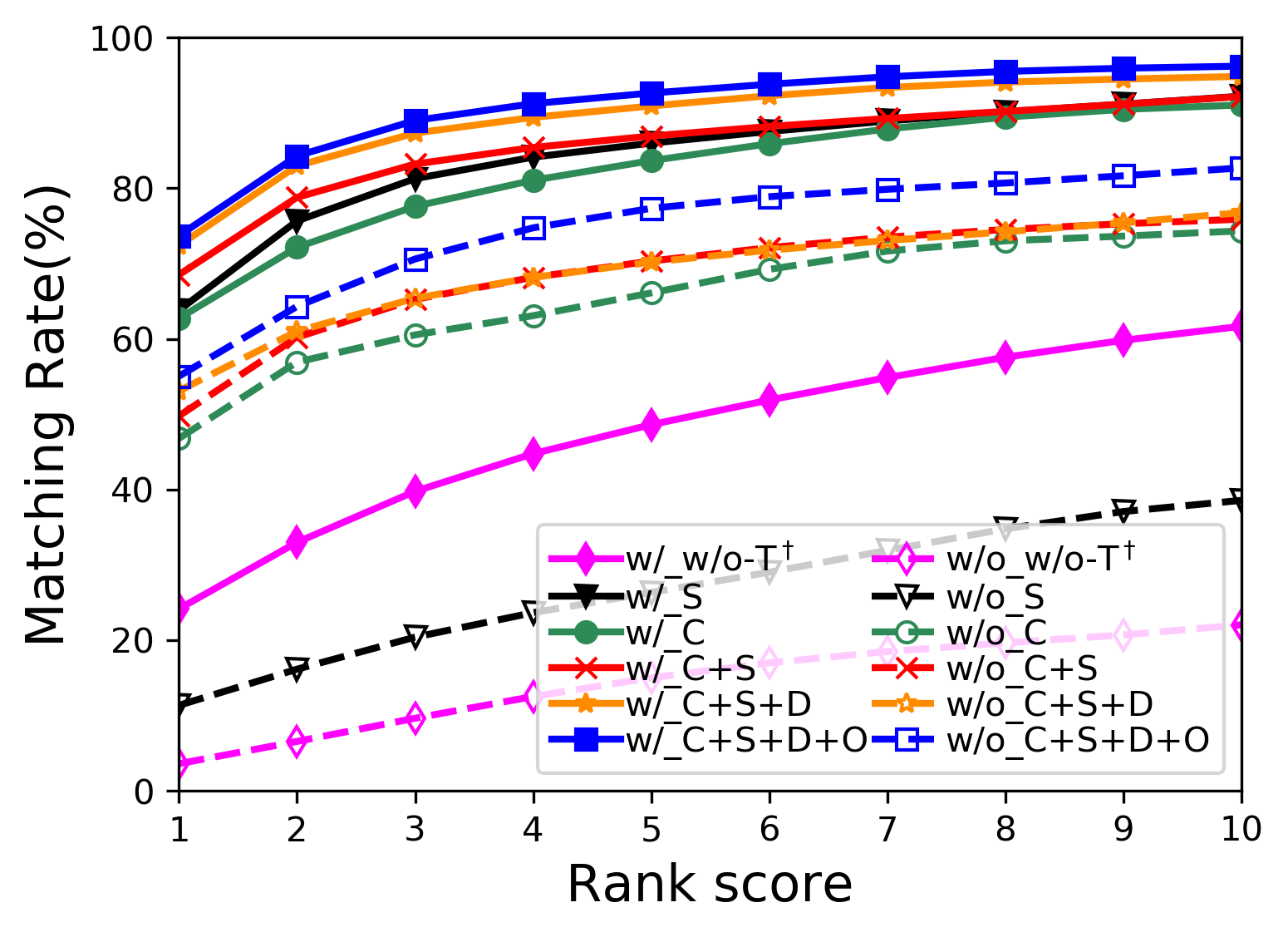}}
        \subfloat[Partial-REID]{\includegraphics[width=0.24\textwidth]{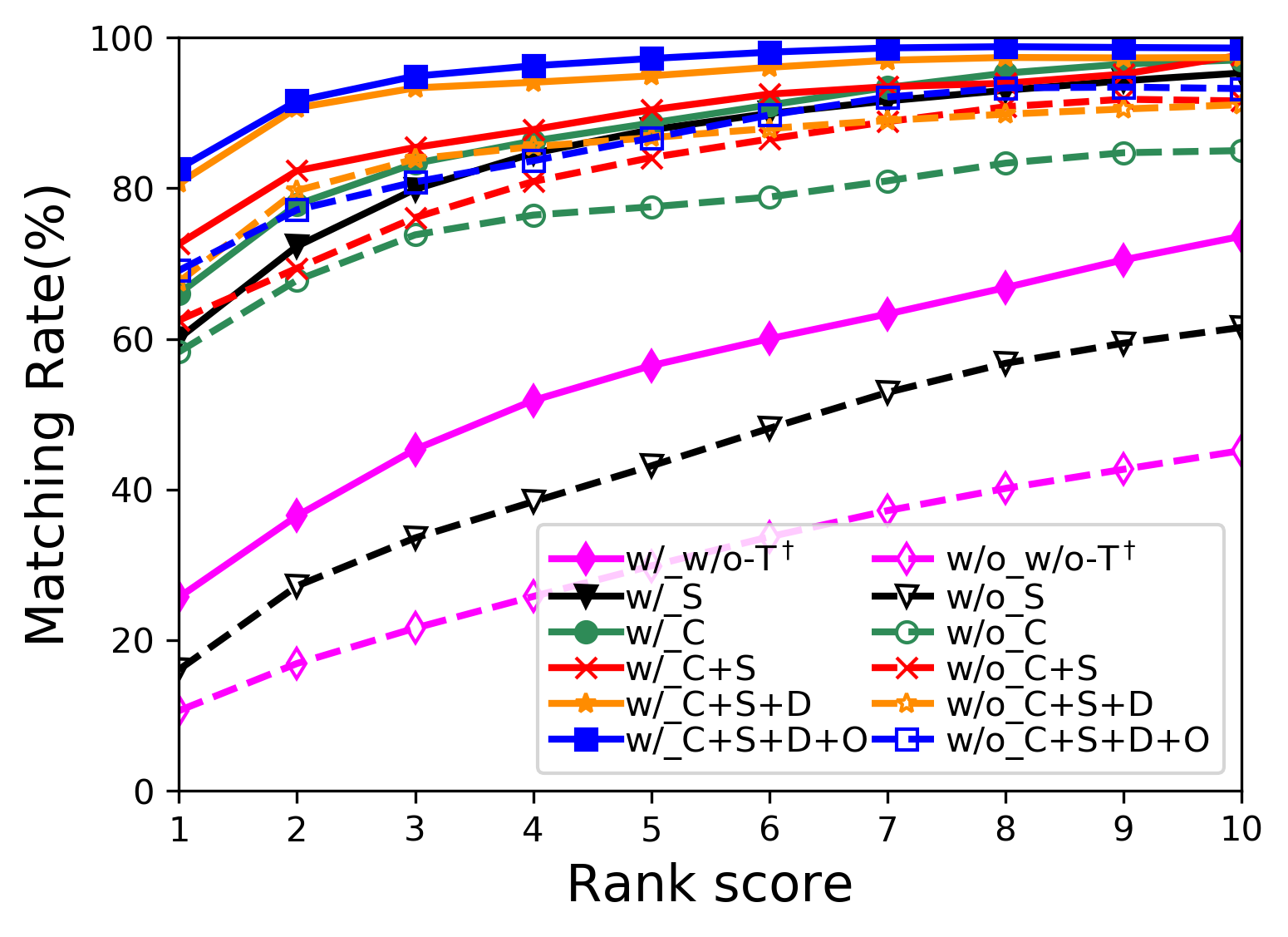}}
        \subfloat[P-DukeMTMC-reID]{\includegraphics[width=0.24\textwidth]{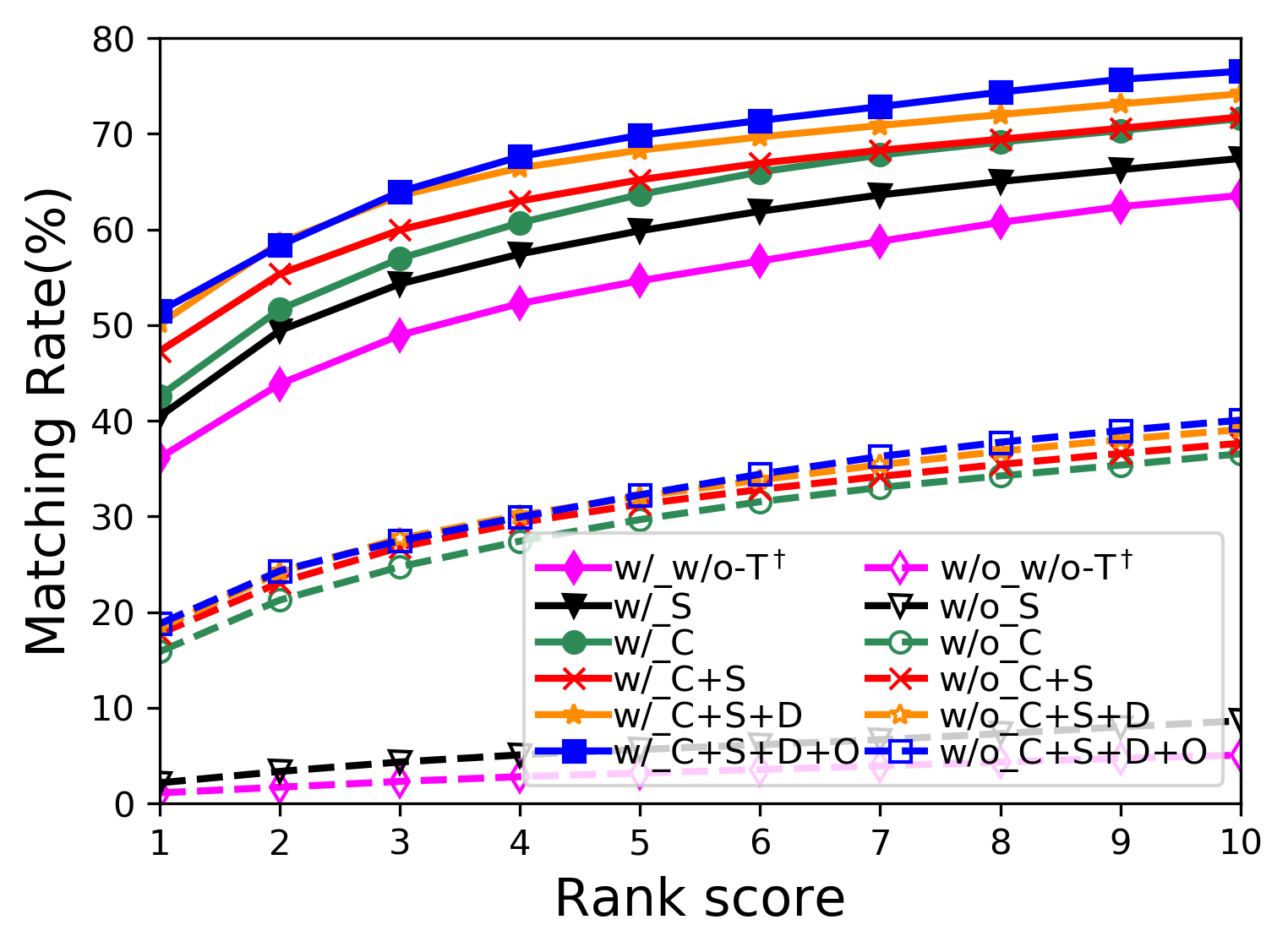}}
		\subfloat[P-ETHZ]{\includegraphics[width=0.24\textwidth]{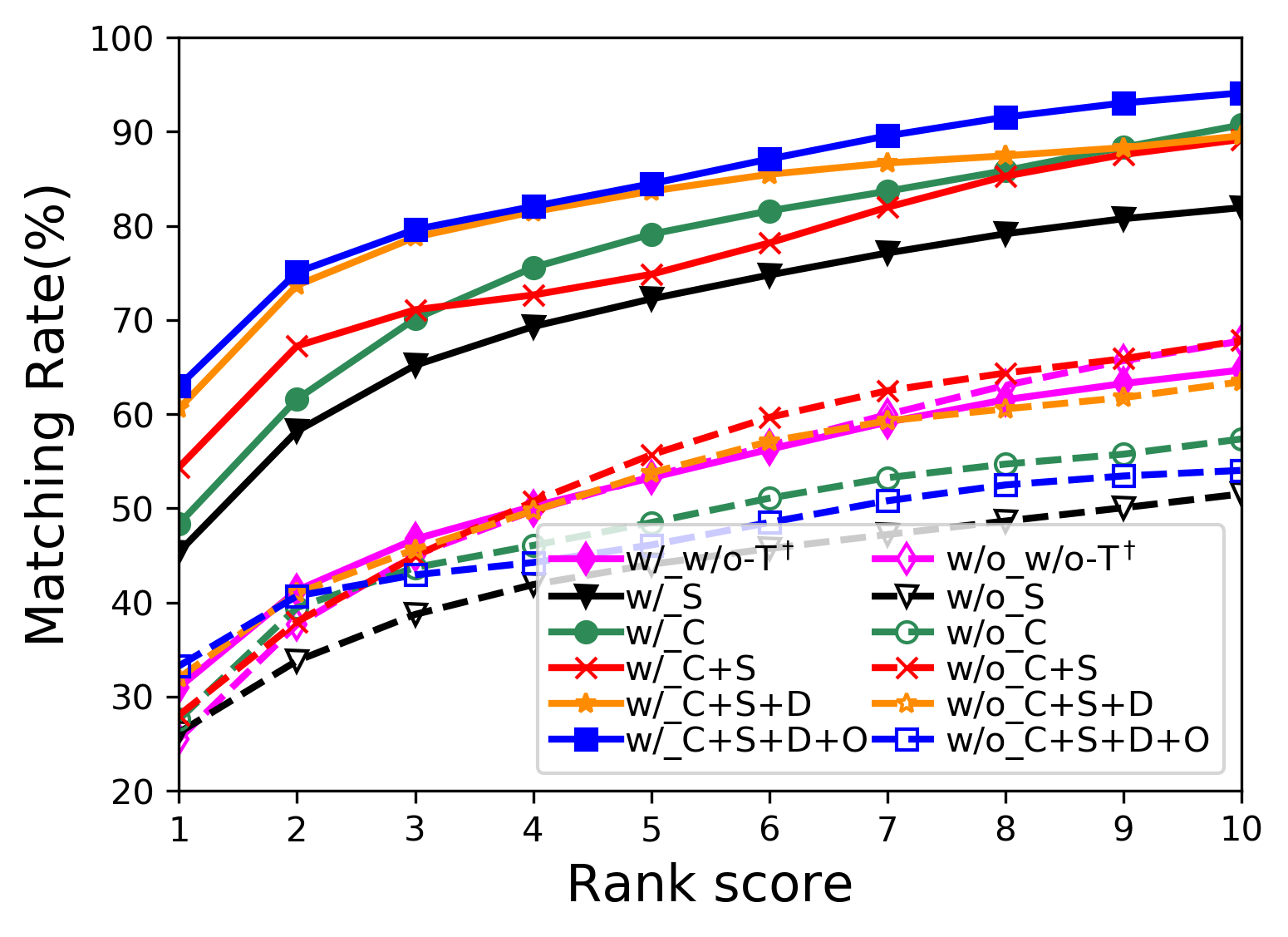}}
		\caption{CMC curves of key components. Dashed lines represent the performances without the "student" stage (The left column of icon is the unsupervised ones) while solid lines represent those with the "student" stage (The right column of icon is the supervised ones).}
		\label{Fig:pic_cmc}
	\end{center}
\end{figure*}

\noindent\textbf{Comparisons of key components.}
To validate the effectiveness of key components in our framework, we conduct the comparisons between different cases on four occluded person re-id datasets, Occluded-REID, Partial-REID, P-DukeMTMC-reID and P-ETHZ, as displayed in Table \ref{Tab:table1} and Figure \ref{Fig:pic_cmc}. The upper form in Table \ref{Tab:table1} is the case without the "student" stage and the lower one is the case with the "student" stage, which donate the unsupervised experiments and the supervised experiments, respectively. As shown in Table \ref{Tab:table1}, there are poor performances in the case without simulation teaching of the "teacher" (w/o-T$^\dagger$) or with the wrong teaching task (T$^\dagger$:S). And the teacher networks equipped with different components lead to different performances. No matter whether in the unsupervised experiments or the supervised experiments, the teacher network with both the classification branch and the co-saliency branch (T$^\dagger$:C+S) performs better than the one only with the classifier (T$^\dagger$:C), which indicates the network make a promotion with the help of the co-saliency branch. In addition, it illustrates the effectiveness of the cross-domain simulator by comparing between the results of the row T$^\dagger$:C+S+D and the row T$^\dagger$:C+S. After using the cross-domain simulator, the performances outperform the previous case by a large margin (improve rank-1 accuracy by 6.53\%, 8.00\%, 3.32\% and 13.5\% without the student "stage", 5.69\%, 11.00\%, 6.00\% and 11.42\% with the student "stage"). Finally, we find that the OBC loss also can bring the network a little upgrade by comparing between the results of the row T$^\dagger$:C+S+D+O (Ours) and the row T$^\dagger$:C+S+D. In summary, there is of great importance to carry out simulation teaching of the "teacher" and key components in our method are able to improve the performance of our framework by degrees.

\begin{figure}[t]
	\begin{center}		
		\includegraphics[width=0.47\textwidth]{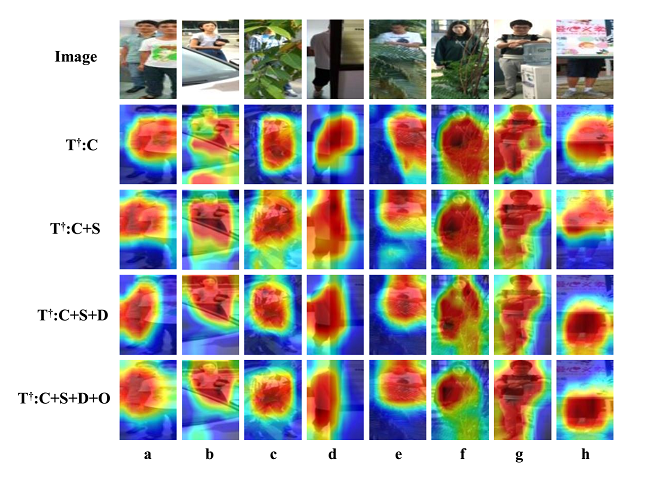}
		\caption{Visualization of saliency maps for real-world occluded person samples.}
		\label{Fig:pic6}
	\end{center}
\end{figure}

Besides, we visualize the saliency maps generated by average pooling all feature maps of the last convolution layer in the backbone from real-world occluded person images. As shown in Figure \ref{Fig:pic6}, the samples include different kinds of occlusions, e.g., dynamic obstacles (a. other person; b. mobile vehicle), environmental vegetations (c. surrounding tree; e. ,f. brushwood) and static obstacles (d. ,h. buildings, g. appliance). It is obviously that saliency maps of T$^\dagger$:C always focuses on the large regions beyond the target person body parts even the other wrong regions. By contrast, saliency maps of T$^\dagger$:C+S+D and T$^\dagger$:C+S pay more attention to the most essential regions, which indicates that key components can boost the performances of our framework effectively. Finally, the visualizations of combining all the key components (T$^\dagger$:C+S+D+O) achieve the best performance.

\begin{figure}[t]
	\begin{center}		
		\includegraphics[width=0.45\textwidth]{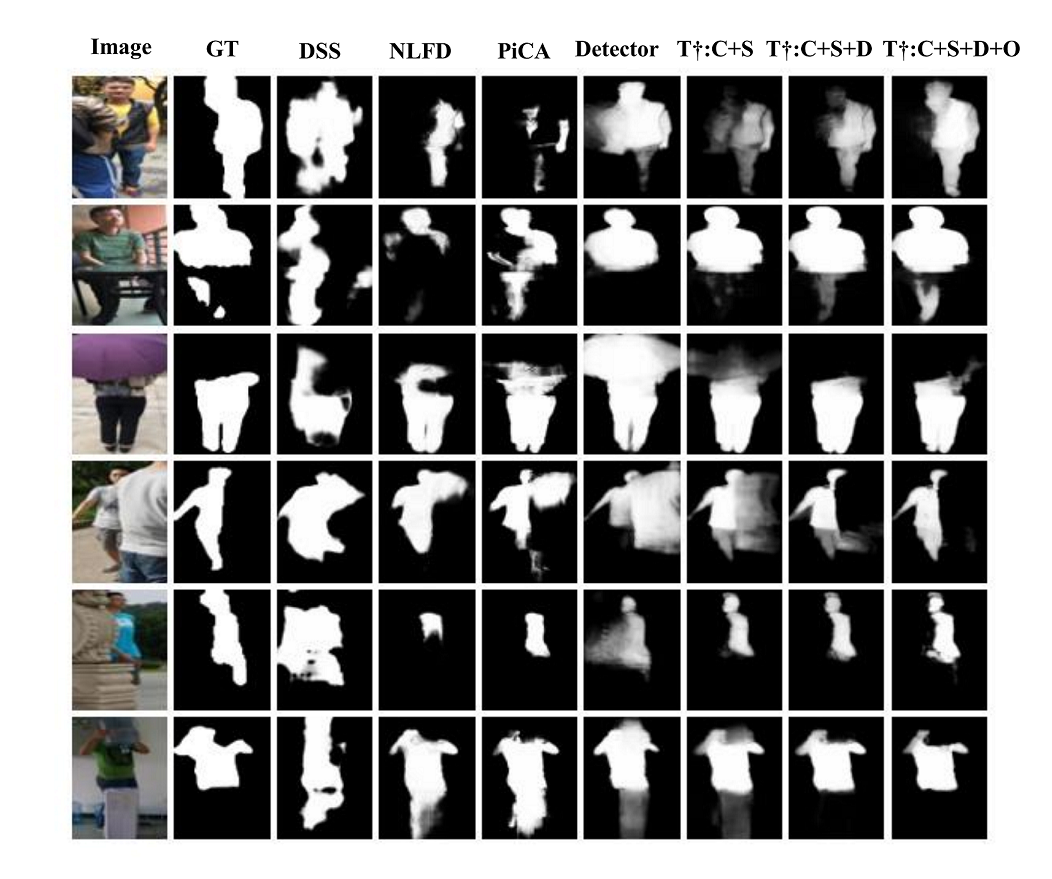}
		\caption{Samples of existing salient object detectors and our co-saliency branch.}
		\label{Fig:pic5_3}
	\end{center}
\end{figure}

\noindent\textbf{Effectiveness of co-saliency branch.} In the "student" stage, we use the co-saliency branch from the "teacher" stage as a salient person detector to get the saliency ground truth of occluded persons. To evaluate the performance of salient occluded person detection, we compare our co-saliency branch with the salient object detector used in the initial stage and 3 existing superior methods in salient object detection, DSS \cite{hou2017deeply}, NLDF \cite{luo2017non} and PiCA \cite{liu2018picanet}. As shown in Figure \ref{Fig:pic5_3}, it can be seen that only the person body parts are more salient and the boundaries between persons and occlusions are clearer using our co-saliency branch (``T$^\dagger$:C+S'' and ``T$^\dagger$:C+S+D''), which is superior than the other contrast methods. For quantitative analysis, the precision, recall and F-measure score are listed in Table \ref{Tab:table2}. Our proposal takes the lead in most metrics, which proves that the proposed network could also improve the salient occluded person detection.

\begin{table}[t]
\begin{center}
\caption{Quantitative comparisons between our co-saliency branch and other salient object detectors.}
\label{Tab:table2}
\footnotesize
\tabcolsep1pt 
\renewcommand\arraystretch{1.2} 
\begin{tabular}{l|p{1cm}<{\centering}p{1cm}<{\centering}p{1cm}<{\centering}|p{1cm}<{\centering}p{1cm}<{\centering}p{1cm}<{\centering}}
\hline
Dataset & \multicolumn{3}{p{3cm}<{\centering}|}{Occluded-REID} & \multicolumn{3}{p{3cm}<{\centering}}{Partial-REID} \\ \hline
Method & Precision & Recall & F-score & Precision & Recall & F-score \\ \hline
DSS \cite{hou2017deeply} & 0.76 & 0.30 & 0.56 & 0.72 & 0.33 & 0.57 \\
NLDF \cite{luo2017non} & 0.79 & 0.52 & 0.71 & 0.73 & 0.28 & 0.53 \\
PiCA \cite{liu2018picanet} & 0.78 & 0.25 & 0.52 & 0.71 & 0.28 & 0.52 \\ \hline
T$^\dagger$:detector & 0.80 & 0.78 & 0.80 & 0.79 & \textcolor{red}{0.46} & \textcolor{red}{0.68} \\
T$^\dagger$:C+S & 0.81 & 0.81 & 0.81 & 0.80 & 0.42 & 0.66 \\
T$^\dagger$:C+S+D & \textcolor{red}{0.83} & 0.80 & \textcolor{red}{0.82} & \textcolor{red}{0.81} & 0.42 & 0.67 \\
T$^\dagger$:C+S+D+O & 0.82 & \textcolor{red}{0.82} & \textcolor{red}{0.82} & \textcolor{red}{0.81} & 0.44 & \textcolor{red}{0.68} \\
\hline
\end{tabular}
\end{center}
\end{table}

\begin{figure}[t]
	\begin{center}		
		\subfloat[Occluded-REID]{\includegraphics[width=0.23\textwidth]{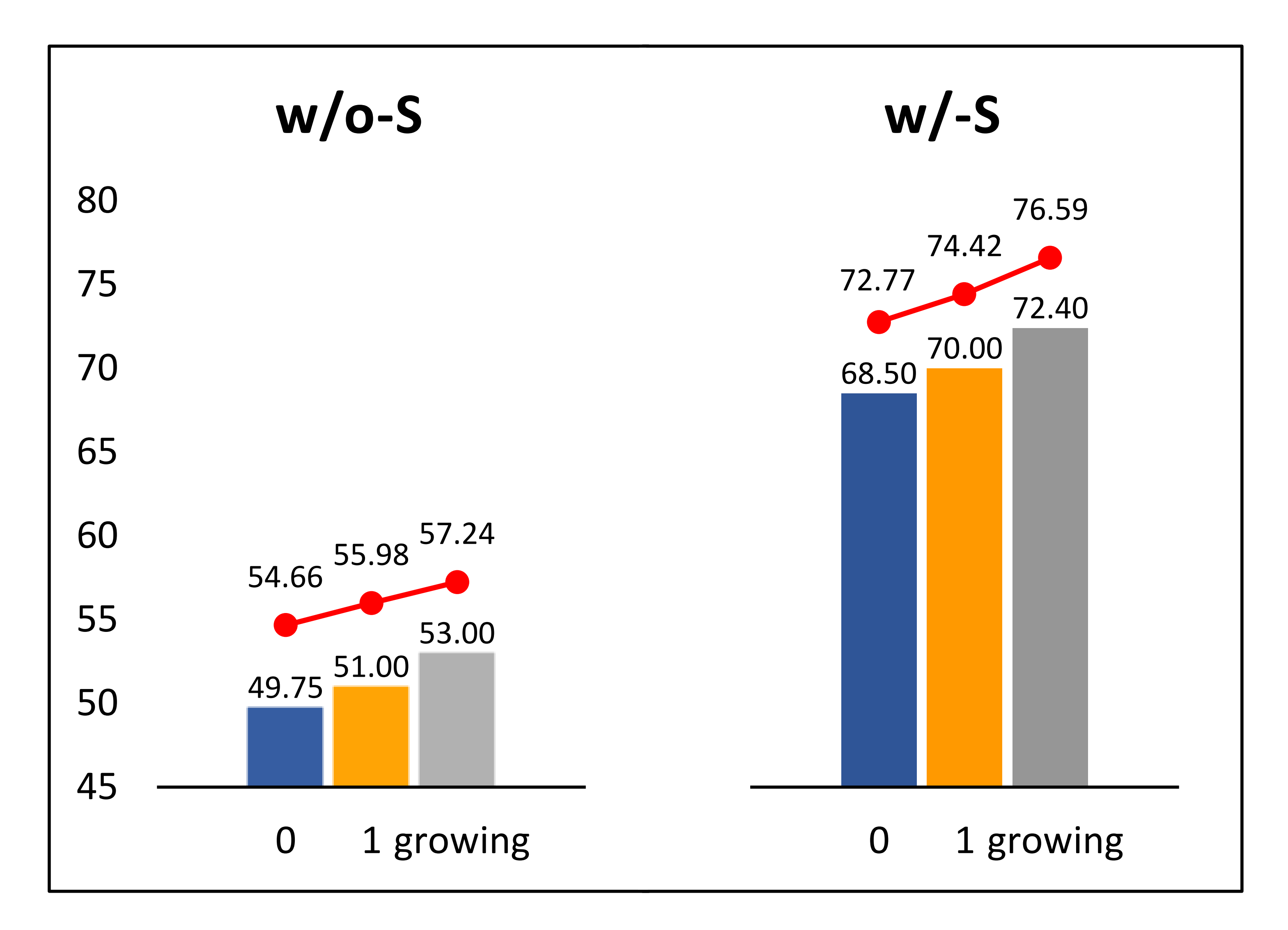}}
        \subfloat[Partial-REID]{\includegraphics[width=0.2263\textwidth]{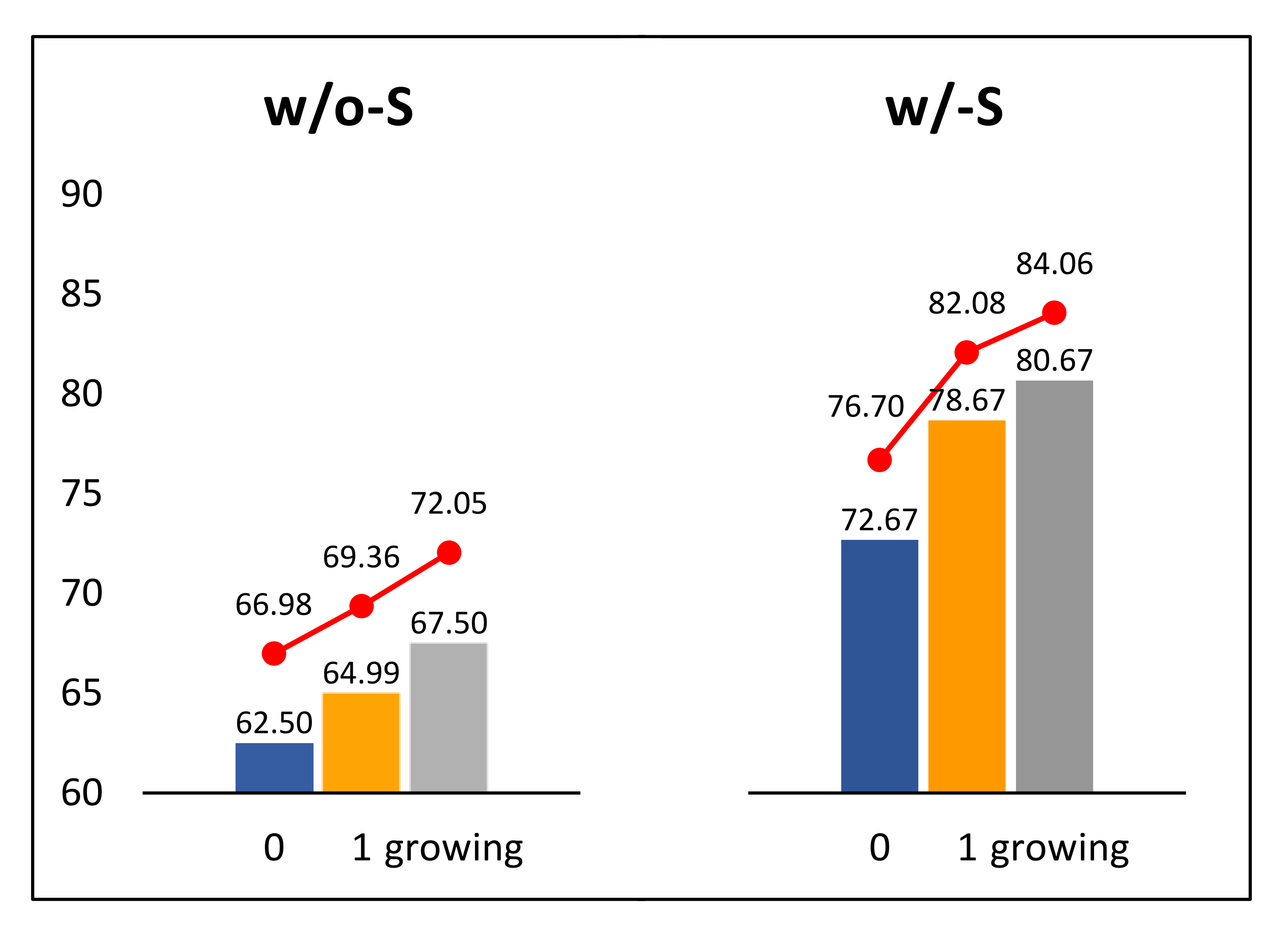}}
		\caption{Performance comparisons of the cross-domain simulators with 0, 1 and growing probabilities, respectively. Bar charts denote rank-1 and line charts denote mAP.}
		\label{Fig:pic_probability}
	\end{center}
\end{figure}

\noindent\textbf{Effectiveness of cross-domain simulator.} The cross-domain simulator transforms full-body person data to simulated occluded person data with a growing probability during training. To further demonstrate that the growing probability of the cross-domain simulator is advantageous, we compare the representations of the cross-domain simulator with the constant probabilities 0, 1 and the growing probability on Occluded-REID and Partial-REID. The constant probability 0 means there are all full-body person data without any transformation. While 1 means there are all simulated occluded person data. As can be seen in Table \ref{Fig:pic_probability} and Figure \ref{Tab:table5}, the cross-domain simulator with the growing probability performs better than those with 0 and 1 in both the unsupervised and supervised cases. This is reasonable because the cross-domain simulator with the growing probability enable the network to learn discriminative and occlusion-robust features via the gradual transformation from full-body person data to simulated occluded person data.

\begin{table}[t]
\caption{Comparisons of the cross-domain simulators with 0, 1 and growing probabilities on rank-1/5 and mAP.}
\label{Tab:table5}
\footnotesize
\tabcolsep4pt 
\renewcommand\arraystretch{1.25} 
\begin{center}
\begin{tabular}{c|l|ccc|ccc}
\hline
 & Dataset & \multicolumn{3}{c|}{Occluded-REID} & \multicolumn{3}{c}{Partial-REID} \\ \hline
 & Probability & r=1 & r=5 & mAP & r=1 & r=5 & mAP \\ \hline
\multirow{3}{*}{w/o-S$^\dagger$} & 0 & 49.75 & 70.50 & 54.66 & 62.50 & 85.00 & 66.98 \\
 & 1 & 51.00 & \textcolor{red}{72.00} & 55.98 & 64.99 & 85.83 & 69.36 \\
 & growing & \textcolor{red}{53.00} & 70.25 & \textcolor{red}{57.24} & \textcolor{red}{67.50} & \textcolor{red}{86.67} & \textcolor{red}{72.05} \\ \hline
\multirow{3}{*}{w/-S$^\dagger$} & 0 & 68.50 & 86.90 & 72.77 & 72.67 & 90.67 & 76.70 \\
 & 1 & 70.00 & 90.60 & 74.42 & 78.67 & 94.33 & 82.08 \\
 & growing & \textcolor{red}{72.40} & \textcolor{red}{90.80} & \textcolor{red}{76.59} & \textcolor{red}{80.67} & \textcolor{red}{94.99} & \textcolor{red}{84.06} \\
\hline
\end{tabular}
\end{center}
\end{table}

\begin{figure*}[t]
	\begin{center}		
		\subfloat[Occluded-REID]{\includegraphics[width=0.24\textwidth]{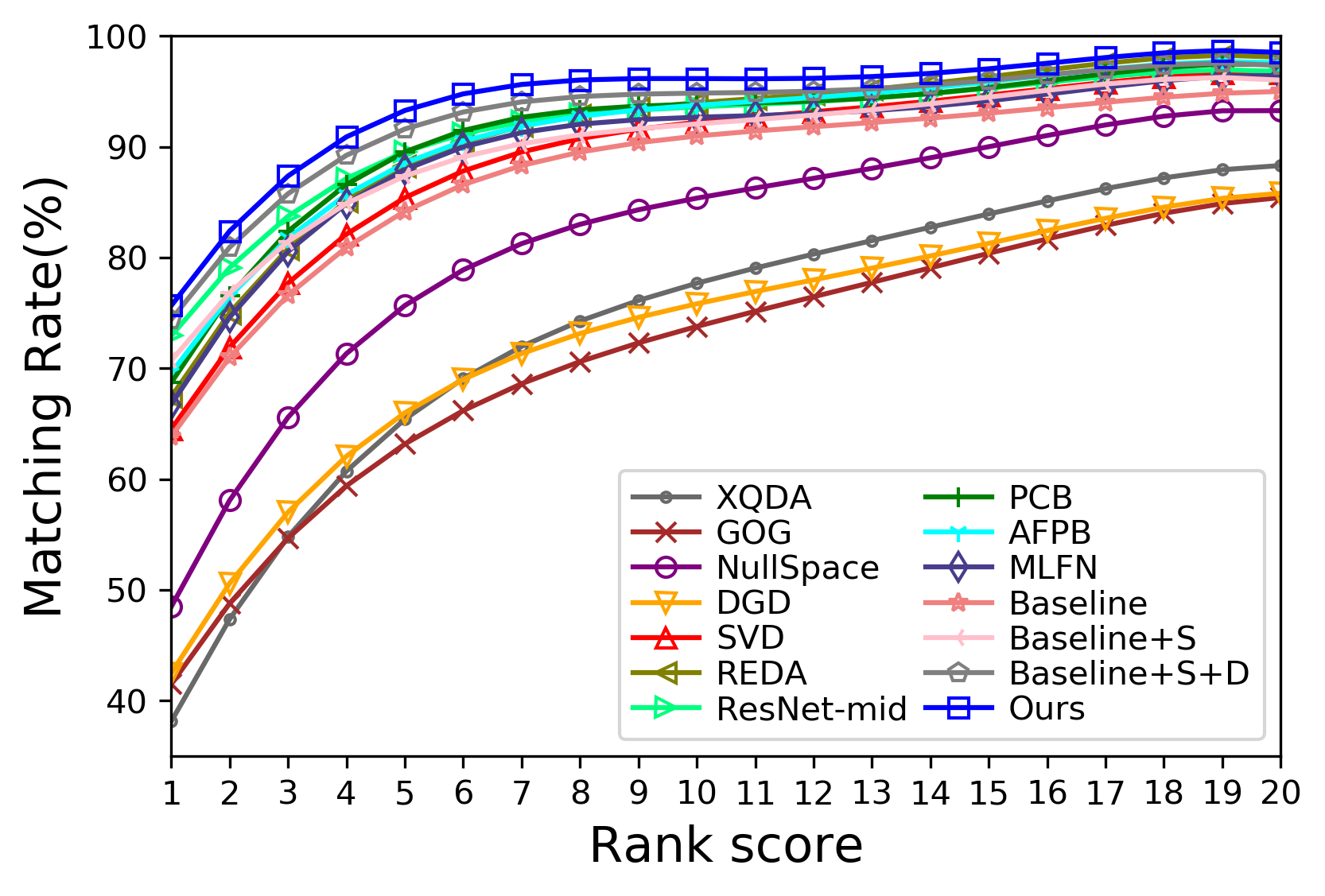}}
        \subfloat[Partial-REID]{\includegraphics[width=0.24\textwidth]{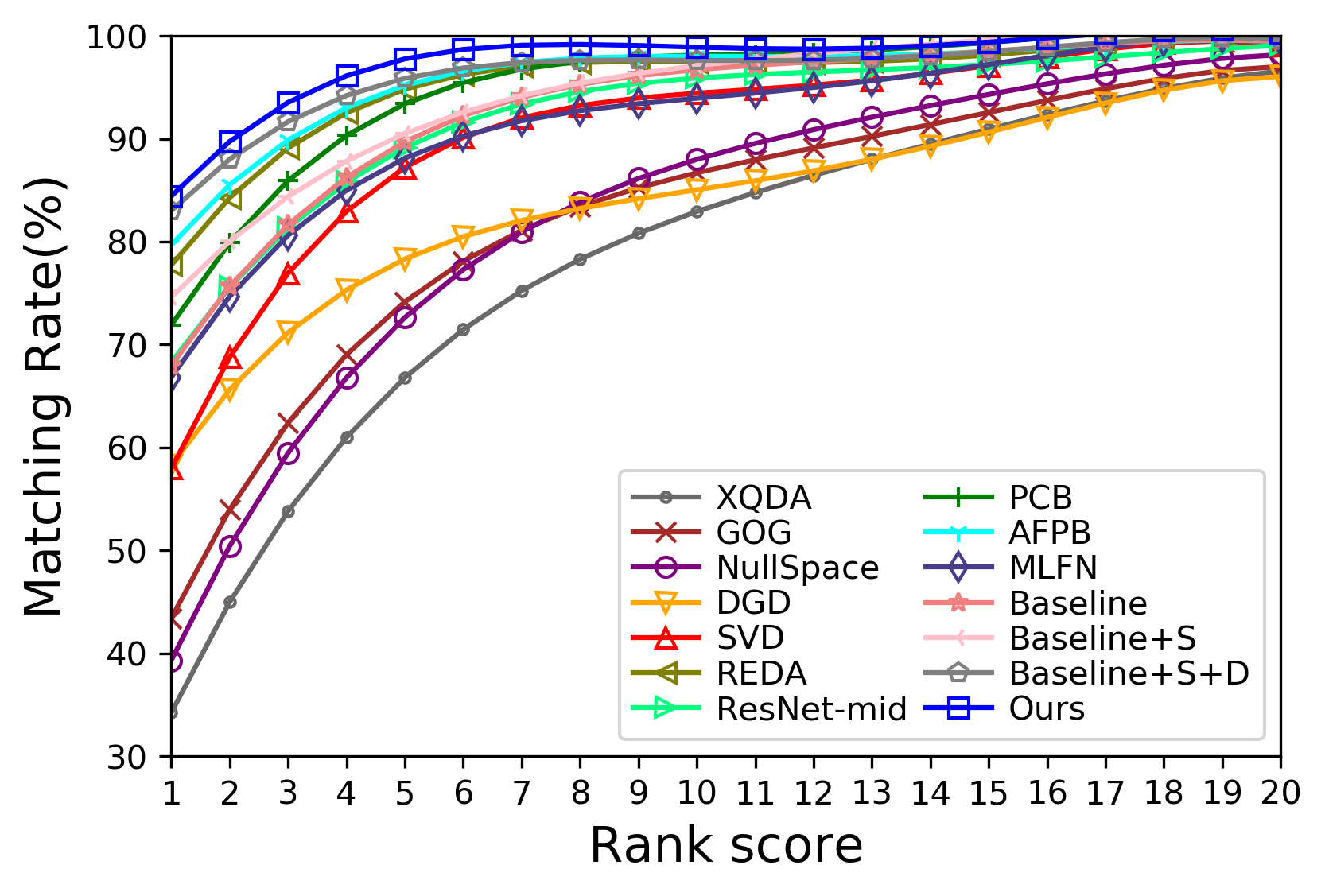}}
        \subfloat[P-DukeMTMC-reID]{\includegraphics[width=0.24\textwidth]{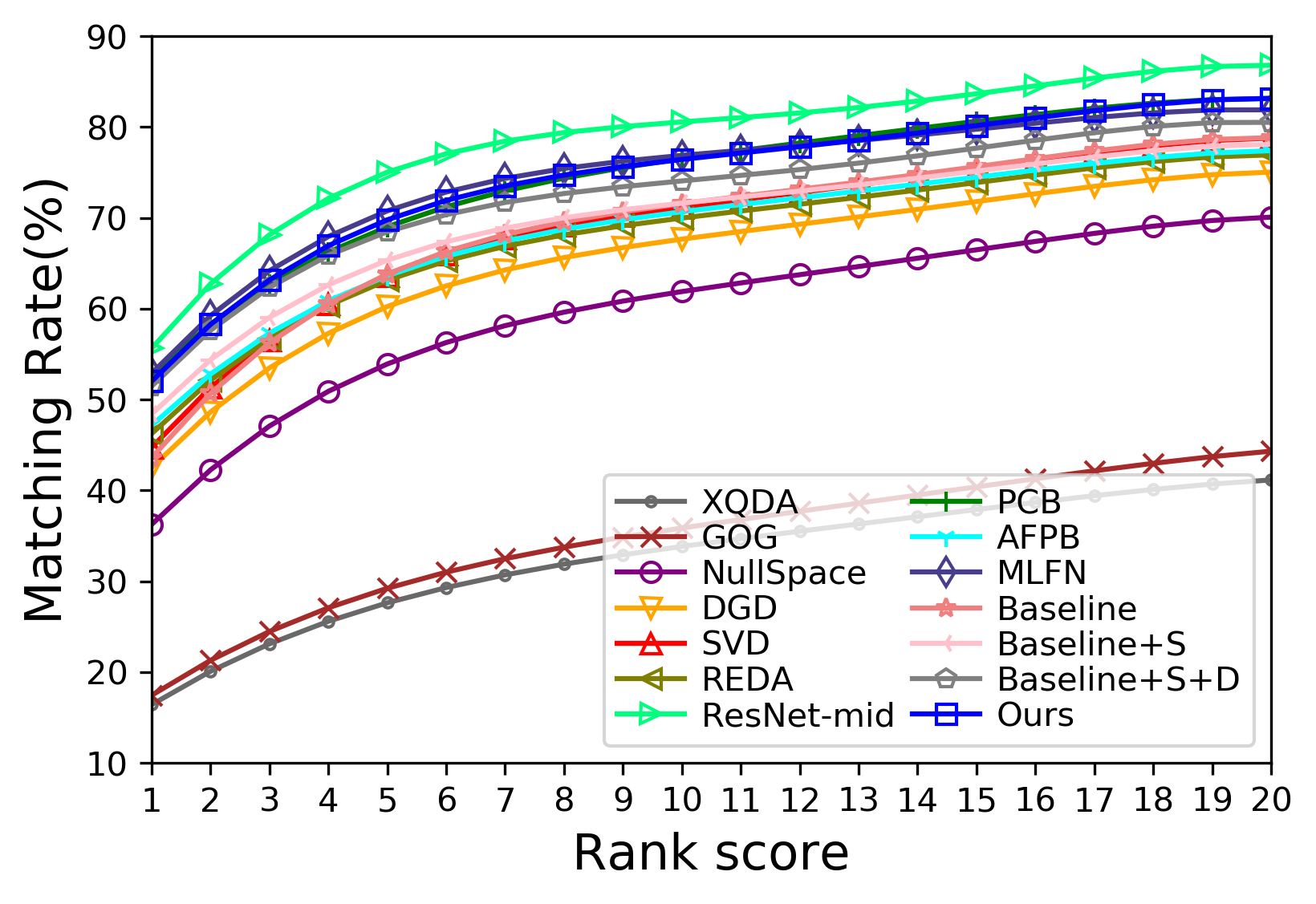}}
		\subfloat[P-ETHZ]{\includegraphics[width=0.24\textwidth]{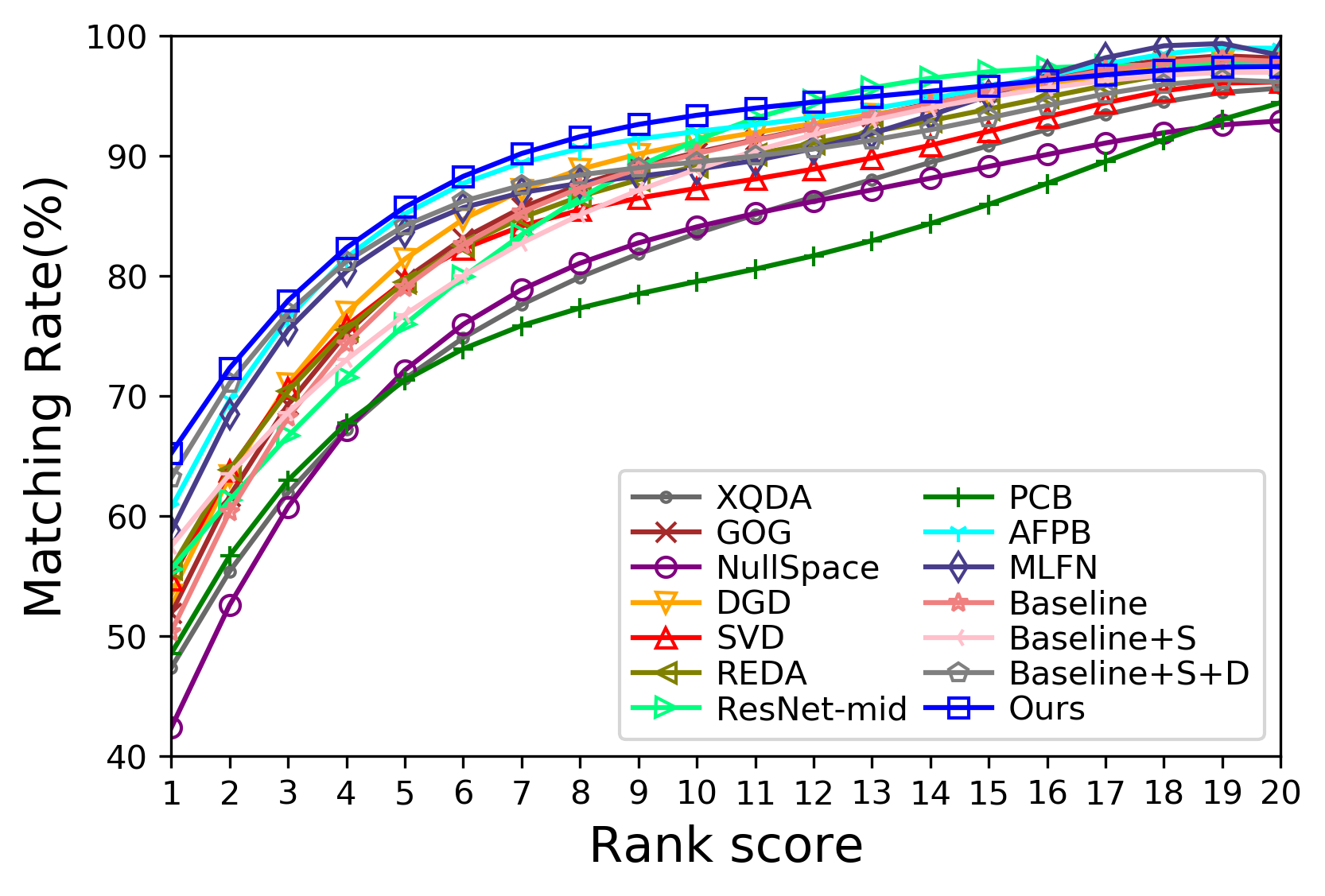}}
		\caption{Comparisons with state-of-the-art on CMC curve.}
		\label{Fig:pic_sota}
	\end{center}
\end{figure*}

\subsection{Comparisons with the State-of-the-art}
In this section, we compare our method with the mainstream occluded person re-id works and other state-of-the-art methods in person re-id on four occluded person re-id datasets, Partial-REID, Occluded-REID, P-DukeMTMC-reID and P-ETHZ. First, We evaluate our proposal in the unsupervised experiment to compare with the recent occluded person re-id works. As listed in Table \ref{Tab:table3}, our method (Ours) on Partial-REID achieves $69.2\%$ on rank-1, which presents the best performance currently. We also compare our methods (Baseline, Baseline+S, Baseline+S+D, Ours) with three traditional methods (the first term, Table \ref{Tab:table4}) and seven state-of-the-art methods based on deep learning (the second term, Table \ref{Tab:table4}) , as shown in Table \ref{Tab:table4} and Figure \ref{Fig:pic_sota}. It is evident that our methods and other deep learning methods are more superior than those traditional methods. Besides, our method takes the first and second places almost in all categories, which show better performances than other state-of-the-art methods. On P-DukeMTMC-reID, our method does not perform the best, because P-DukeMTMC-reID is a large-scare occluded person dataset while our method is more effective for the small-scale occluded person dataset. In general, our method is of great superiority on occluded person re-id problem.

\begin{table}[!htbp]
\begin{center}
\caption{Comparisons with mainstream occluded person re-id methods on Partial-REID.}
\label{Tab:table3}
\footnotesize
\tabcolsep3pt 
\renewcommand\arraystretch{1.2} 
\begin{tabular}{l|c|ccc}
\hline
Dataset & \multirow{2}{*}{Type} & \multicolumn{3}{c}{Partial-REID} \\
Method & & r = 1 & r = 5 & r = 10 \\ \hline
SWM \cite{Zheng2016Partial}& supervised & 24.4 & 52.3 & 61.3 \\
AMC \cite{Zheng2016Partial}& supervised & 33.3 & 52.0 & 62.0 \\
AMC+SWM \cite{Zheng2016Partial} & supervised & 36.0 & 60.0 & 70.7 \\ \hline
DSR(Multi-scale) \cite{DBLP:conf/cvpr/HeLLS18} & unsupervised & 43.0 & 75.0 & 76.7 \\
AFPB \cite{Zhuo2018Occluded} & unsupervised & 51.7 & \textcolor{green}{79.2}  & \textcolor{green}{86.7} \\
SCPNet-baseline \cite{DBLP:journals/corr/abs-1810-06996} & unsupervised & \textcolor{green}{60.0} & 78.3 & 83.7 \\
SCPNet-a \cite{DBLP:journals/corr/abs-1810-06996} & unsupervised & \textcolor{blue}{68.3} & \textcolor{blue}{80.7} & \textcolor{blue}{88.3} \\
Ours & unsupervised & \textcolor{red}{69.2} & \textcolor{red}{85.8} & \textcolor{red}{93.3} \\
\hline
\end{tabular}
\end{center}
\end{table}

\begin{table}[!htbp]
\begin{center}
\caption{Comparisons with state-of-the-art on rank-1/5.}
\label{Tab:table4}
\footnotesize
\tabcolsep0.62pt 
\renewcommand\arraystretch{1.2} 
\begin{tabular}{l|p{0.77cm}<{\centering}p{0.77cm}<{\centering}|p{0.75cm}<{\centering}p{0.75cm}<{\centering}|p{0.75cm}<{\centering}p{0.75cm}<{\centering}|p{0.75cm}<{\centering}p{0.7cm}<{\centering}}	
\hline  
\multirow{1}{*}{\rotatebox{0}{Dataset}} & \multicolumn{2}{p{1.55cm}<{\centering}|}{Occluded-REID} & \multicolumn{2}{p{1.5cm}<{\centering}|}{Partial-REID} & \multicolumn{2}{p{1.5cm}<{\centering}|}{P-DukeMTMC} & \multicolumn{2}{p{1.45cm}<{\centering}}{P-ETHZ} \\
\hline
\multirow{1}{*}{\rotatebox{0}{Methods}} & r=1 & r=5 & r=1 & r=5 & r=1 & r=5 & r=1 & r=5 \\	
\hline
XQDA \cite{liao2015person} & 36.71 &	65.11 &	33.14 &	66.18 &	15.93 &	27.50 & 44.98 & 70.88 \\
GOG \cite{matsukawa2016hierarchical} & 40.50 & 63.16 & 41.92 & 74.00 & 17.10 & 29.27 & 49.17 & 79.29 \\
NullSpace \cite{zhang2016learning} & 46.47 & 75.36 & 37.73 & 72.12 & 35.17 & 53.65 & 40.16 & 71.53 \\		
\hline
DGD \cite{xiao2016learning} & 41.43 & 65.74 & 56.83 & 77.70 & 41.53 & 60.09 & 51.23 & 81.01 \\
SVDNet \cite{sun2017svdnet} & 63.13 & 85.13 & 56.05 & 87.06 & 43.47 & 63.41 & 52.21 & 78.95 \\
REDA \cite{zhong2017random} & 65.79 & 87.88 & 76.19 & 94.57 & 45.18 & 62.88 & 54.43 & 79.09 \\
ResNet-mid \cite{Qian2017The} & \textcolor{green}{70.80} & 88.90 & 66.00 & 88.33 & \textcolor{red}{54.89} & \textcolor{red}{75.24} & 54.76 & 71.43 \\
PCB \cite{Sun2018Beyond} & 66.60 & \textcolor{green}{89.19} & 69.99 & 93.67 & \textcolor{green}{51.42} & 68.77 & 45.24 & 69.05 \\
AFPB \cite{Zhuo2018Occluded} & 68.14 & 88.29 & \textcolor{green}{78.52} & \textcolor{green}{94.87} & 46.15 & 63.47 & \textcolor{green}{58.15} & \textcolor{blue}{84.61} \\
MLFN \cite{chang2018multi} & 64.70 & 87.70 & 64.33 & 87.33 & 50.95 & \textcolor{green}{70.34} & 57.14 & 83.33 \\
\hline	
Baseline \cite{he2016deep} & 62.70 &	83.30 &	66.00 & 87.99 & 42.58 & 63.75 & 48.33 & 79.05 \\
Baseline+S & 68.50 & 86.90 & 72.67 & 90.67 & 47.27 & 65.20 & 54.28 & 75.24 \\
Baseline+S+D & \textcolor{blue}{72.40} & \textcolor{blue}{90.80} & \textcolor{blue}{80.67} & \textcolor{blue}{94.99} & 50.11 & 68.40 & \textcolor{blue}{60.48} & \textcolor{green}{83.81} \\
Ours & \textcolor{red}{73.67} & \textcolor{red}{92.87} & \textcolor{red}{82.67} & \textcolor{red}{97.00} & \textcolor{blue}{51.42} & \textcolor{blue}{69.72} & \textcolor{red}{62.86} & \textcolor{red}{85.24} \\

\hline 
\end{tabular}
\end{center}
\end{table}

\section{Conclusions}

In this work, we propose a teacher-student learning framework for occluded person re-identification. To address the limitation of inadequate occluded person data, the teacher network makes use of large-scale full-body person data to simulate the occluded person re-id. Supported by the co-saliency network and the cross-domain simulator, the teacher network trains a basic model for the student network. The student network then trains a more occlusion-robust model on real-world occluded person data. Experimental results on four public occluded person re-id datasets demonstrate the effectiveness and superiority of our framework.

\bibliographystyle{ACM-Reference-Format}
\bibliography{acmart}

\end{document}